# Topology and geometry of data manifold in deep learning

German Magai
Faculty of computer science
HSE University
gera128a@gmail.com

Anton Ayzenberg
Faculty of computer science
HSE University
ayzenberga@gmail.com

## Abstract

*Despite significant advances in the field of deep learning in applications to various fields, explaining the inner processes of deep learning models remains an important and open question. The purpose of this article is to describe and substantiate the geometric and topological view of the learning process of neural networks. Our attention is focused on the internal representation of neural networks and on the dynamics of changes in the topology and geometry of the data manifold on different layers. We also propose a method for assessing the generalizing ability of neural networks based on topological descriptors. In this paper, we use the concepts of topological data analysis and intrinsic dimension, and we present a wide range of experiments on different datasets and different configurations of convolutional neural network architectures. In addition, we consider the issue of the geometry of adversarial attacks in the classification task and spoofing attacks on face recognition systems. Our work is a contribution to the development of an important area of explainable and interpretable AI through the example of computer vision.*

## 1. Introduction

For many years, deep neural networks (DNN) have shown successful results in applications to many areas: natural and social science, computer vision, text processing and generation, audio, and much more. Despite obvious empirical progress, some deep learning issues are still poorly understood theoretically. It remains an open question why they approximate complex data dependencies so well. One of the main problems of deep learning (DL) models is interpretability, and finding solutions to this problem will provide a principal approach to the design and development of new architectures. In this regard, one of the relevant and most promising topics in the field of DL is the study of the models themselves, the features of their internal structure, behavior on various sets of real data.

In many computer vision tasks, neural networks give better results than humans. There is a hypothesis according to which the classification and categorization problem is reduced to untangling linearly inseparable object manifolds into squeezed, compact representations that can be separated by a linear hyperplane [18, 70]. This assumption applies to both neural networks and brain models and also consistent with findings from neuroscience [71].

Formulating topological and geometrical-based theoretical framework for deep learning can significantly improve the theoretical understanding of many phenomena in this field and be an important step for explainable AI [1].

In this paper, we propose an interpretation based on geometry and topological data analysis (TDA) of the learning process of neural networks, their ability to generalize and performance in classification problems. And also we considered the phenomenon of adversarial and spoofing attacks on DNN models from the point of view of TDA and intrinsic dimension (ID). Our study is the first to systematically consider the topological properties of data in the internal representation across all depth of convolutional neural networks (CNN). We address the following issues:

• The dynamics of changes in the topological descriptors of data manifolds in the internal representation in the learning process on the example of different architectures of DNNs, activation functions and datasets.

• Connection of topological descriptors with the generalization ability of DNNs.

• Geometry and topology of adversarial attacks and spoofing manifold on face recognition systems.

## 2. Related works

Approaches based on algebraic topology and geometry can give a new vision of the above problems related to understanding DL algorithms, and the search for an original, effective solution to them by formulating a descriptive model of learning processes and the structure of neural networks. A set of such tools and methods is generalized in TDA. The available results can be divided into 3 groups: topological and geometric aspects of learning and performance of DNN, investigation of the internal representation of DNN and geometric interpretation of adversarial attacks. Some articles aim to clarify the processes that take place inside DNN during training and inference, while others suggest improving existing DL solutions.

**Topological and geometric aspects of learning and performance of DNN:** Particular attention is paid to the analysis of the generalizability, capacity and expressiveness of deep learning models using topology and geometry methods. Gus in [2] raises the question of whether the learning and generalizability of a DNNs depends on the homologically complexity of a particular dataset. On the other hand, authors of [3] and [4] investigate the performance of DNNs depending on the ID of the training dataset and notice that the generalization error does not depend on the extrinsic dimension of data. In [5] it is proved that a DNNs can successfully learn with help of SGD to solve a binary classification problem when a number of conditions on the width, length and sample complexity are met with certain geometric properties of the training dataset. There is also a series of works that consider the expressive power of DNNs through the analysis of the topology of the decision boundary [6-8]. In some works, a computational graph of a neural network is considered [9-10]. In [77] proved that the Hausdorff dimension of the SGD trajectory can be related to generalization error. [78] consider the $PH_{dim}$ of the learning trajectory in the parameter space, in the optimization SGD process, this is the first work that connects $PH_{dim}$ and theoretical problems of deep learning. [11] analyzing the space of weights in convolution filters, the authors come to the conclusion that the topological features of this space change during the learning of CNN.

**Geometry of adversarial attacks:** In [12] adversarial attacks are detected by analyzing the computational graph. In [13,14], the influence of local dimension on adversarial disturbance is analyzed. [15] consider the curvature of the decisive boundary and associate it with the vulnerability of the DNN to adversarial examples. [16] Uses persistent homology to detect various types of image distortion. In [64] manifold-based interpretation is proposed.

**Internal representation of neural networks:** To understand the learning process of DNNs, it is important to have an idea of the dynamics and changes in geometric properties within the space of embeddings on different layers. The paper [17] argues that the learning process of DNNs is associated with the untangling of object manifolds, on which the data lie, when passing through the layers of the DNN, and a quantitative assessment of the untangling is proposed. [18] analyze the change in the geometry of object manifolds (dimension, radius, capacity) during training and argue that as a result of the evolution of object manifolds in a well-trained DNN, by the end of the hierarchy of layers, manifolds become linearly separable. [19] focus on the problem of memorization and generalization when training DNN and come to the conclusion that memorization occurs on deeper layers due to a decrease in the radius and size of object manifolds. [20] also consider transformations of data in a DNN and attempts to formulate deep learning theory in terms of Riemannian geometry. In [67] the ability of DNN to learn an efficient low-dimensional representation is considered, and [68] deals with the DNN-based transformation of data into minimal representation. [79] use persistence landscapes to analyze the dynamics of topological complexity across all layers of the neural network, according to their results, topological descriptors are not always simplified in the learning process. In many ways, our work is a continuation and improvement of the ideas outlined in [21], where it is argued that when passing through the layers of a FC neural network, the data topology is simplified and the Betti numbers are reduced, and the ReLU activation function contributes to better performance. But this work did not use the entire rich arsenal of persistent homologies and calculated the Betti numbers for a fixed epsilon. In addition, it is important for us [22], which shows the dynamics of ID when passing data on different layers throughout the depth of the model and its relationship with performance and generalization gap using the example of several modern CNNs architectures, the authors conclude that neural networks transform data into low-dimensional representations. [23] shown that clusters with different density peaks are formed on different layers, which reflects the semantic hierarchy of the ImageNet.

## 3. Topological descriptors and persistent homology dimension

The geometric and topological properties of the manifold on which the data lie can be viewed from different points of view: from the intrinsic and the extrinsic. Intrinsic characteristics describe the local perspective: distance measured on the manifold, tangent vectors, angles between them, curvature, normal vector. In contrast, concepts such as the shape of a manifold and dimension are extrinsic. Also, recently, approaches have been actively developed in which the problems and tasks of deep learning, use the methods of computational geometry and applied algebraic topology. In this section, we present a definition and brief description of the concepts that we will use to explore data manifolds in DNNs.

### 3.1. Persistent homology and barcodes

The main tool for topological data analysis is the concept of persistent homology. This construction summarizes the process of changing the Betti numbers during the filtration. The basic concepts and a brief theoretical introduction to the field of TDA will be given below. Details can be found in [24 - 26].

To be able to work with objects in vector space X, it is necessary to construct simplicial complexes K from them. We use the Vietoris-Rips complex K(X; r), which is constructed from simplexes that emerge from the pairwise intersection of balls with radius r. The properties of simplicial complex K are expressed in homology groups.

$C_i(K, F)$ is a free abelian group of i-dimensional simplexes with coefficients in a field F. $\partial$ is the operator of taking a boundary called simplicial differential, a chain complex is a sequence of abelian groups of simplicial chains and simplicial homomorphisms between them:

$$\ldots C_3(K,F) \xrightarrow{\partial_3} C_2(K,F) \xrightarrow{\partial_2} C_1(K,F) \xrightarrow{\partial_1} C_0(K,F) \xrightarrow{\partial_0} 0 \quad (1)$$

Then one can define the group $Z_i(K) = \text{Ker }\partial_i = \{C \in C_i(K) \mid \partial_i C = 0\}$ of i-dimensional cycles. The subgroup $B_i(K) = \text{Im }\partial_{i+1} = \{C \in C_i(K) \mid \exists b \in C_{i+1}(K): C = \partial_{i+1} b\}$ of i-dimensional boundaries. The group $H_i(K,F) = Z_i(K,F)/B_i(K,F)$ is the i-dimensional simplicial homology group. These are i-dim cycles minus (with trivialization) i-dim boundaries. $\beta_i(K) = \dim H_i(K,F)$ is called the i-e Betti number, it shows the number of connected components, cycles and cavities for dimension 0, 1, 2 respectively. An increasing sequence of simplicial complexes is called a filtration: $\{K_i\}_{i \in \mathbb{Z}_{\geq 0}} = K_0 \subseteq K_1 \subseteq K_2 \cdots K_s$ (Fig 1).

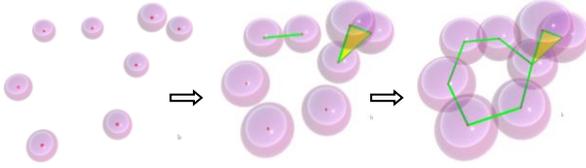

Figure 1: Geometric realization of Vietoris-Rips complexes in filtration, green denotes 1-simplexes, yellow 2-simplexes.

The change of homological features of simplicial complexes in filtration over time can be expressed in the form of a barcode – a set of lifetimes intervals (Fig. 2). At the moment of time $t_{birth}^n$, a n-th homology is born, and at the moment $t_{death}^n$ it dies, that is, it disappears from the filtration. $I_n = t_{death}^n - t_{birth}^n$ the lifetime of the n-th homology feature, called lifespan, it is obvious that it cannot be negative.

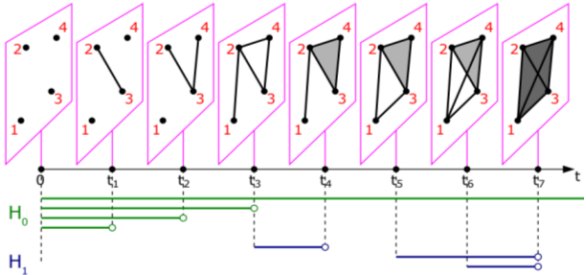

Figure 2: An example of a barcode with discrete time $t_i$, green color denotes homology of dimension 0, blue – 1.

A long interval in barcode means that we have a sufficiently long-lived n-homology, which is a persistent topological feature, it helps to evaluate the topological properties of a space. An equivalent representation of the barcode is the persistent diagram. This is a plane with coordinates of the birth $t_{birth}$ and death $t_{death}$ times, each homology is indicated by a point with a color corresponding to the grading. All points are above the diagonal, and the further away the point is from diagonal, the more significant is the homological feature. To analyze the topological properties of dataset X, a topological descriptor summarizing [29] persistent diagrams and barcodes will be used.

The power-weighted sum of N lifespans for the i-th homology dimension is denoted as follows:

$$E_\alpha^i(X) = \sum_{n=1}^{N} I_n^\alpha \quad (2)$$

where $\alpha \geq 0$, it is noted in [27] that $E_1^0(X_n)$ it is equal to the length of Euclidean minimal spanning tree of set of n points in $\mathbb{R}^d$. Finding a geometric interpretation of $E_\alpha^1(X_n)$ could be a topic for future research. Numerical experiments and calculations of persistent diagrams were carried out using Ripser [28]. In the barcode, there is always one topological feature with an infinite lifetime, we do not take it into account. These characteristics of the persistent diagram give a lot of important topological information.

### 3.2. Persistent homology fractal dimension

We are increasingly faced with very high-dimensional data in computer vision problems, sound signal processing, in the analysis of gene expression, and others. According to the manifold hypothesis [30], the data X lies on a low-dimensional submanifold: $X \subseteq M^n \subseteq \mathbb{R}^d$, where d – extrinsic dimension, n – ID. [31] Proved that given sufficient sample complexity, we can model a low-dimensional representation of the data with some error.

It is possible to divide the methods for estimating the ID into two classes [32, 33]:1) Local methods estimate the dimension at each data point from its local neighborhood, and then calculate the average over the local estimates of the ID: NN algorithm [34], MLE [35], nonlinear manifold learning methods. 2) Global methods estimate the dimension using the entire dataset, assuming that the dataset has the same dimension throughout: PCA, MDS, k-NNG, GMST [36]. This group also includes fractal-based methods: Hausdorff dimension, correlation dimension and box-counting dimension.

If we consider the persistent diagrams obtained from a set of points in $\mathbb{R}^d$, it is often possible to notice an accumulation of homological features in the diagonal area and near zero, they are born and die quickly. These points are considered to be topological noise that does not carry useful information. However, the decay rate of "noise" depends on the size of the submanifold on which the set of points lies.

[39] introduces the concept of persistent homological fractal dimension, generalizing Steele's result [38] for any homology group dimensions and based on Adams [76].

Let X be a bounded subset of a metric space and μ a measure defined on X. For each i ∈ N define the persistent homology fractal dimension of μ:

$$\text{PHdim}_\alpha^i(\mu) = \frac{\alpha}{1-\beta} \quad (3)$$

where

$$\beta = \limsup_{n\to\infty} \frac{\log\left(\mathbb{E}\left(E_\alpha^i(x_1,..x_n)\right)\right)}{\log(n)} \quad (4)$$

and $x_1,..,x_n$ are sampled independently from μ. That is, $\text{PHdim}_\alpha^i(\mu) = d$ if $E_\alpha^i(x_1,..x_n)$ scales as $n^{\frac{d-\alpha}{d}}$ and $\alpha \geq 0$. Larger values of α give relatively more weight to large intervals than to small ones. In other words, the persistent homological dimension can be estimated by analyzing the asymptotic behavior at n→∞ of $E_\alpha^i(x_1,..x_n)$ for any i. Schweinhart in [39] proved that if μ satisfies the hypothesis of Ahlfors regularity, then $\text{PHdim}_\alpha^0(\mu)$ equals the Hausdorff dimension of the support of μ. See Appendix A.

In practice, the $\text{PHdim}_\alpha^i(\mu)$ can be calculated as follows: for real $\{m_1, m_2..m_k\}$ logarithmic values $\log_{10}(n)=m_i$, we randomly sample n points from X and calculate the $E_\alpha^i(x_1,..x_n)$. We then use linear regression to fitting the power law for obtained values n and $E_\alpha^i(x_1,..x_n)$ [37]. In all experiments in our work, the calculations were carried out for the case i = 0 and α = 1, and the $\text{PHdim}_1^0(\mu)$ will be denoted $\text{PH}_{dim}$. For the case of low-dim. space, you can use the Boruvka minimum spanning tree algorithm [40].

Unfortunately, the problem of existing algorithmic methods for calculating the ID is that for a correct estimate, an exponentially large amount of data is needed, which leads to a systematic error and a significant underestimation of the obtained ID [33]. To assess the performance of the method for calculating the $\text{PH}_{dim}$, we use the approach outlined in [4]: generate synthetic data using a BigGAN [65], to set the necessary *d* upper bound for the ID, we fix 128 - *d* elements of the hidden vector equal to zero. In table 1 we compare the accuracy of $\text{PH}_{dim}$ with other ID estimation approaches: Correlation dimension [74], TwoNN [34], MLE [35].[75] was used to implement other methods.

## 4. Topology and geometry of data manifold

To develop new, high-performance CNNs architectures and solve open problems in the field of deep learning, it is necessary to understand the essence of the processes that occur during training, as well as what affects the expressive power and ability of deep learning models to generalization.

DNN learns an internal low-dimensional representation of the data manifold on different layers and DNN training can be interpreted in terms of evolution of embedding manifold representations, changes in topology and data geometry.

| ID\real ID | 8 | 16 | 24 | 32 | 48 | 64 | 96 |
|---|---|---|---|---|---|---|---|
| **PHdim$_0$** | **8.96** | **17.72** | **23.49** | **30** | **32** | **43.2** | **46.58** |
| TwoNN | 10 | 20 | 24 | 28 | 30 | 38 | 36 |
| MLE | 11 | 19 | 23 | 26 | 28 | 33 | 28 |
| CorrDim | 10.17 | 12.97 | 16.47 | 17.34 | 18.4 | 15.8 | 14.05 |

Table 1: Estimation of the synthetic datasets ID. See Appendix C.

An embedding manifold $X^n \subseteq \mathbb{R}^d$ is a low-dimensional representation of the object data manifold within a DNN, where d - the width of the layer. In the process of passing the data manifold from layer to layer along the all depth, the manifold's characteristics: shape, ID, curvature [69], etc. change. Here we track the change in topological descriptors and $\text{PH}_{dim}$ at different depths of model and associate them with the success of training in image classification tasks and with generalization error. We do a large set of experiments, considering a wide range of CNNs with different architectures and activation functions. The results are also checked against several datasets. Next, we analyzing the topological descriptors of the adversarial manifold and investigate the $\text{PH}_{dim}$ of spoofing attacks.

### 4.1. Evolution of topological descriptors and PH$_{dim}$ inside neural networks

The design ideology of modern CNNs architectures is based on the principle of structural blocks (moduls). The architecture consists of specific blocks connected in series, which include a set of different operations, such as 1D, 2D and depthwise convolution, pooling, skip-connections, compression, batchnormalization, etc. Different configurations and combinations of these operations give different efficiency, accuracy, and computational complexity and at the same time affect differently the change in the geometry of the object manifold.

A deep neural network is a function DNN: $\mathbb{R}^{in} \to \mathbb{R}^{out}$ defined by the composition DNN = softmax ∘ $f_d$ ∘…∘$f_2$ ∘ $f_1$, where $f_i$ - is a function of the i-th structural block, softmax – output activation function and $M_i^d=f_i(M_{i-1}^n)$ manifold generated as a result of the action of this block. In the case of CNN, on each layer we have generated feature maps as a result of the convolution operation with a step s and the size of the kernel h × w. At different depths, CNNs extract features corresponding to different levels of abstraction; on the first layers, features correspond to the image style and general geometric primitives, and at the end – to high level specific features that separate some classes from others.

We analyze the embedding representation at the output of each structural block and we get this representation using the global average pooling. One of the main results of this section is the demonstration of the behavior of the topological descriptors of the data embedding manifold on different layers during training at different epochs, (Fig. 3). The experiments were carried out on the example of several modern architectures: ResNet [41], SE-ResNet [42], MobileNetV2 [43], VGG [44].

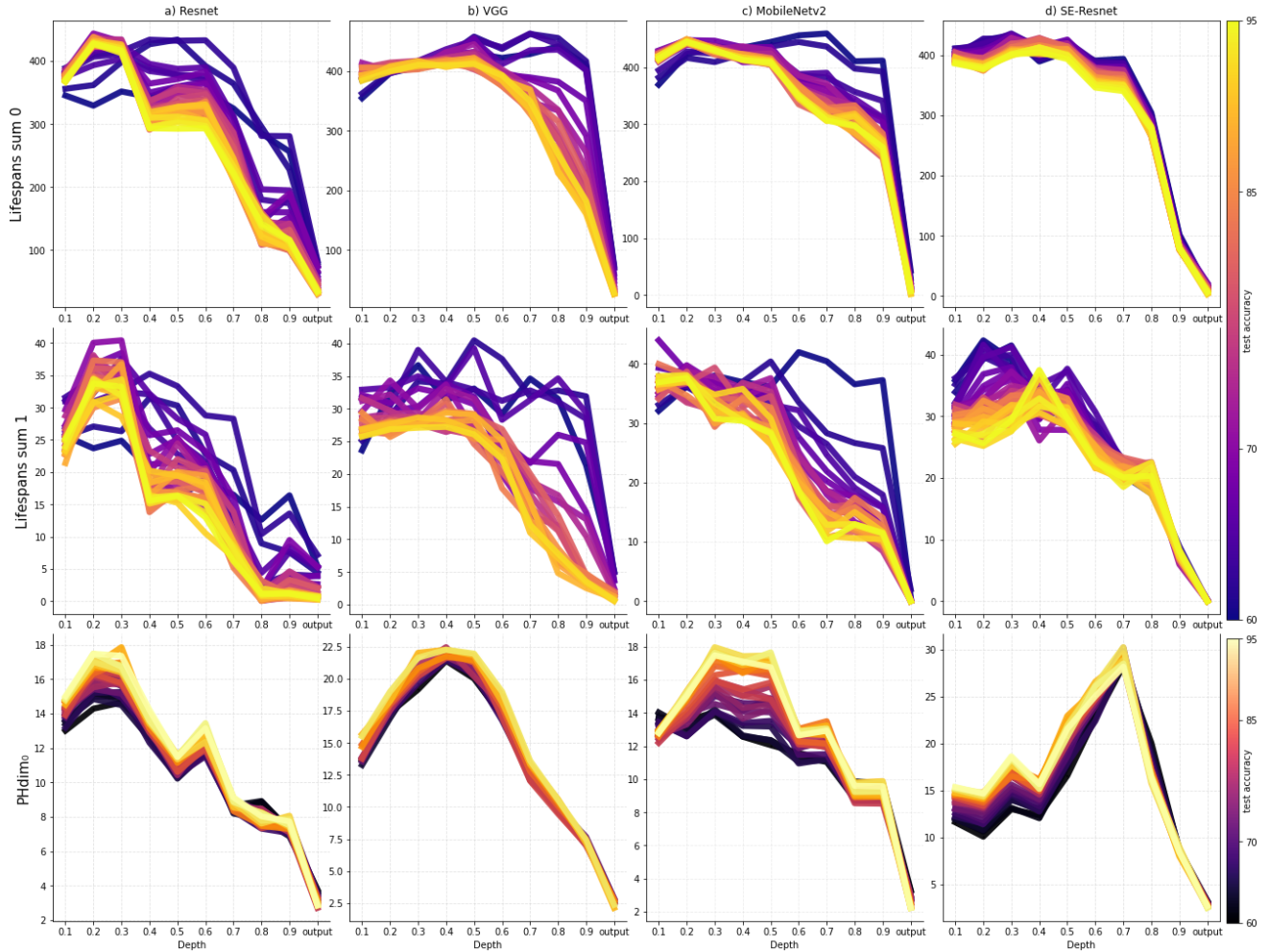

Figure 3: Change in topological descriptors and PH$_{dim}$ (y-axis) on different blocks of neural networks during training at different epochs, x-axis – relative depth from 0.1 (first layers) to 1 (last output fully connected layer).

Experiments are done using the CIFAR-10. For experiments, we feed a randomly selected batches of 300 examples from train dataset of each class to the input of the CNNs several times, and then average the results across batches and over all 10 classes. To eliminate the influence of the ambient space, all models have a fixed width 128 of the input and output of blocks. According to the observation of the evolution of the topology of a train dataset manifold when passing through a CNNs, in the process of training a neural network to learn more efficiently and faster to lower the values of topological descriptors at the all depth. On different architectures of CNN, the evolution of the data manifolds occurs in different ways, this may be a consequence of the unique arrangement of structural modules (blocks) that underlie a particular architecture.

In the case of a ResNet, the dynamics of a decrease in topological descriptors shows a decrease from the very first blocks, and in the case of MobileNetV2 with linear bottleneck and SE-ResNet with squeeze-and-excitation block, the topology is simplified from the middle of the layer hierarchy. In the process of training on different layers, the data form clear clusters that correspond to their classes, and as shown in this section, the shape is simplified, expressed in the reduction of topological descriptors.

The theoretical substantiation of the change in PH$_{dim}$ (Fig. 3) is due to the assumption that in an ideal well-trained neural network in the feature space at the output, untangled and separated clusters corresponding to classes will have an PH$_{dim}$ = 0, it is natural to expect that a well-trained network will reduce the PH$_{dim}$ of the entire dataset in internal representation, which is consistent with observations in Appendix B.

Different datasets are subject to evolution in different ways within the representation of ResNet model. Fig 4a, 4b shows the difference in data dynamics in the same architecture, but with different datasets: CIFAR-10 [45], Street View House Numbers (SVHN) [46], synthetic data generated based on 10 classes of buildings from the ImageNet.

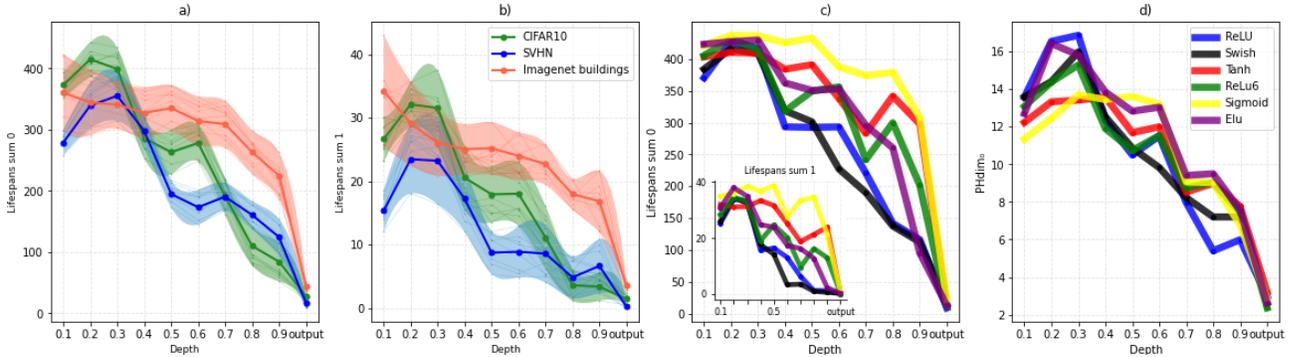

Figure 4: Changing the topological descriptors and $PH_{dim}$ inside ResNet: a, b) Different datasets, thin lines indicate different classes. c) Influence of activation functions on changing topological descriptors. d): Influence of activation functions on changing $PH_{dim}$.

We evaluate the change in topological descriptors and $PH_{dim}$ on intermediate ResNet-101 layers representation with different activation functions. As you can see in the Figure 4c, 4d, different activation functions make different contributions to the topology and $PH_{dim}$ change; with the Swish activation function (black line), model has better accuracy on the test dataset in our experiments, and according to [48] Swish improves CNN performance over ReLU and Sigmoid, helping to alleviate the vanishing gradient problem. The ReLU, ReLU-6 [43] and ELU activation functions are more successful in practice than Tanh and Sigmoid, which is consistent with the dynamics of topological simplification of the train dataset manifold in our experiments. Success of DNNs with ReLU is explained by they are not homeomorphic map [21].

## 4.2. Influence of topology on the generalization ability of CNNs

Neural network generalizations are the ability to learn from a training set to work efficiently on data that was not in the training set. The generalization measure in classification problems is determined by the concept of generalization gap – the difference in the accuracy metric between the indicators on the test and train datasets. But there are also other measures that show the ability of a DNN to generalize, and the development of a sufficiently effective metric from an empirical point of view, while theoretically justified, is one of the most important problems in the statistical learning theory. In addition to classical measures of DNNs expressivity, such as VC-dimension [66] or Rademacher complexity [74], there are many approaches to estimating generalization error [9, 22, 49 - 52].

As shown in Section 4.1, the learning of CNN affects the dynamics of topological descriptors throughout the depth. If we consider the last layer before the softmax activation as the final representation $X^{out}$ of CNN, then we can see the relationship between $E_1^0(X^{out})$ and the learning epoch (Fig. 5). To calculate $E_1^0(X^{out})$, we use 350 examples from each classes and then average over the number of classes.

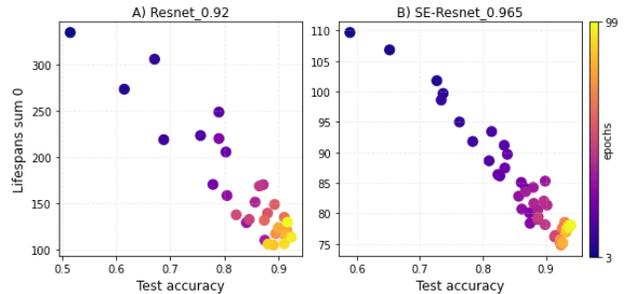

Figure 5: Dependence between $E_0^1(X)$, obtained from the intrinsic representation X of the last layer and the accuracy on the test set in the learning process at different epochs. a) ResNet, r = -0.92. b) SE-ResNet, r = -0.965.

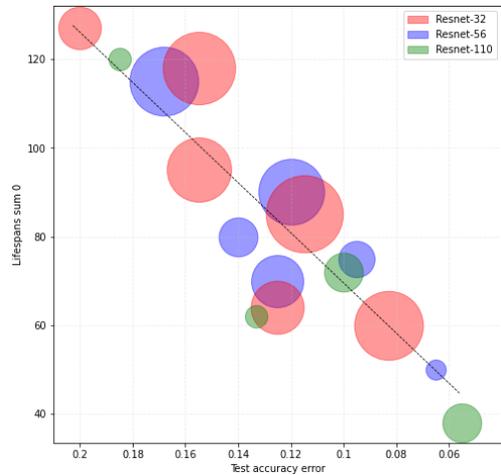

Figure 6: The relationship between model accuracy on a test dataset and topological descriptors in internal representation.

To estimate the generalization error, it is necessary to consider not one model at different learning epochs, but a family of trained models with different accuracy on the test data. If the learning process can be understood as a simplification of the topology of the data manifold across all depth, then the models with the simplest data on the last layer $X^{out}$ have the best performance in test dataset.

Our hypothesis is supported and a statistically significant inverse correlation r = -0.9 (Pearson's correlation coefficient) can be observed between the topological description $E_1^0(X^{out})$ and the test accuracy error, and this can be a reliable predictor of CNN generalization in the classification problem. To use a topological predictor, only the train dataset is required without the need for a test data, which is important in the case of limited data. For experimental verification, a dataset of trained 50 models of the family of architectures ResNet-32, ResNet-56, ResNet-110 with different learning hyperparameters was formed: regularization, learning rate, weight decay, batch size, with or without augmentation. All models are trained on the CIFAR-10 dataset to 100% train accuracy. The Figure 6 shows the dependence of topology on generalization, architectures within one family form clusters indicated by circles with different colors (several models inside one circle). See Appendix B for additional experiments.

### 4.3. Adversarial and spoofing manifolds

Computer vision systems based on neural networks can be susceptible to attacks both in the digital domain, when we have access directly to the input of the model, and in physical domains, when information is supplied to the input of the model through video surveillance. For example, in the task of detecting road signs, it is possible to add adversarial disturbance ε to road signs, which will lead to a detection error. In face recognition systems, spoofing examples [54] are a special type of attack. There are 3 main types of spoofing attacks: a person's mask, a printed face, a static image or video on the display (Fig. 10 right) [59, 60, 61]. As methods for creating adversarial attacks [55] and spoofing attacks evolve, methods for defending against them are also being developed.

To create more effective protection methods, it is necessary to understand all aspects of attack production mechanisms. In this section, we will consider the topology and geometry of adversarial manifold in the classification problem and spoofing examples manifold in biometric identification problems.

An adversarial attack is an algorithm that results in the formation of an adversarial image. For example, let $x \in \mathbb{R}^d$ be an image, where $d = h \times w \times c$ is the dimension of an image with c channels. The DNN F: $\mathbb{R}^d \to \{1...k\}$ performs the k-classification task, then an adversarial example $x' \in \mathbb{R}^d$ can be created such that: $F(x') \neq F(x)$, while there is a measure of the difference ε between the original image and the adversarial example: $\|x' - x\| \leq \varepsilon$.

Set $X' = \{x'_1, x'_2, ..x'_n\}$ we call adversarial manifold, and we assume that adversarial examples lie on a low-dimensional manifold. We also show the evolution of data manifold in the internal MobileNetV2 representation across all layers in the classification task using our synthetic dataset (Fig. 10 left).

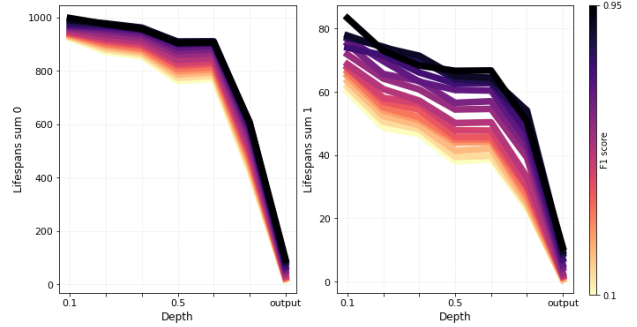

Figure 7: The dynamics of topological characteristics throughout the entire depth of the MobileNetV2 model, depending on the success of the attack (F1-score). Left: $E_1^0(X')$. Right: $E_1^1(X')$.

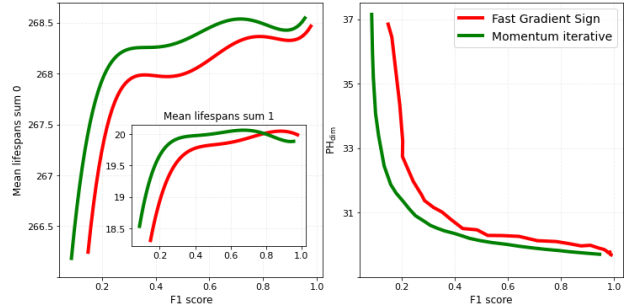

Figure 8: Left: The dynamics of topological characteristics throughout the entire depth of the MobileNetV2, depending on the success of the attack (F1-score). Right: $PH_{dim}$.

FGSM [56] and MI-FGSM [57] were chosen as methods for creating attacks, they were carried out using the library [58]. Depending on the success of the attacks, the model makes more errors and its F1-score decreases. By changing epsilon, we achieve more successful attacks and lower model accuracy. Adversarial examples of all classes in the internal representation of the CNN are concentrated around the decision boundaries corresponding to the target class of the attack, the more successful this is, the denser the clusters and the simpler the topology. This is reflected in a gradual decrease in topological descriptors $E_1^0(X')$ and $E_1^1(X')$ across all layers, depending on the success of the attack (Fig.7). And also the topology of the adversarial manifold (outside the neural network) is simplified, depending on the success (Fig. 8 left). The success of an attack directly affects the topology and dimension of the manifold generated by the attacks. The adversarial noise increases $PH_{dim}$, which would make the attack succeed by slightly increasing the dimension through noise, (Fig. 8 right). In this experiment, when calculating the topological descriptors of the dataset itself, not inside DNNs representation, here we use $E_1^0(X')$ и $E_1^1(X')$ averaged over the number of lifespans in the barcode. See Appendix B.

In the case of spoofing attacks, the person's face is used as a biometric identifier. In order to deceive the system, instead of a real person, a display surface or a photo with his face appears in front of the video capture device.

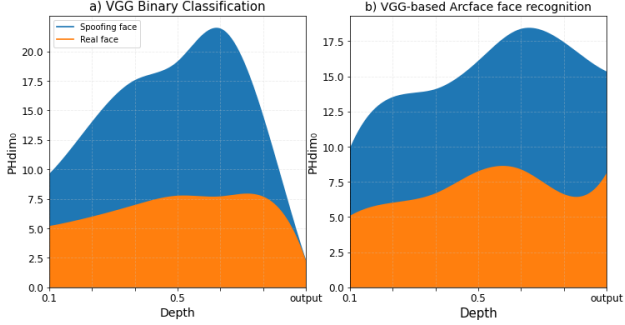

Figure 9: The dynamics of PH$_{dim}$ face anti-spoofing manifold. a) Binary classification. b) Arcface face recognition.

We show here that the PH$_{dim}$ of the real and spoofing faces dataset [59] in the internal representation of the DNNs changes as it passing through all the layers (Fig 9). In the binary classification problem (spoof/real, 2000 examples per each class), the dynamics of the manifold repeats the shape, as in the Section 4.1. In the problem of 10 persons face recognition, the VGG with fixed 32 width architecture is used with a loss function Arcface [62], in the output space the objects lie on a 32-dimensional hypersphere, which simplifies the comparison of objects of the same class with each other. PH$_{dim}$ in this case does not correspond to the shape of the hump. Examples of spoofing attacks have a large dimension due to the presence of effects associated with glare, texture, lighting and other effects. An estimate of the PH$_{dim}$ can be useful to determine the quality and diversity of a spoofing examples dataset. Tests have shown that the CelebA-Spoof dataset PH$_{dim}$ = 14, while the screen attack dataset from [59] PH$_{dim}$ = 10. This can be explained by that CelebA-Spoof presents a wider range of spoof attack types, not just screen attacks Geometrical and topological interpretation of adversarial and spoofing attacks can develop and improve existing attack detection methods.

## 5. Experimental Settings

### 5.1. Neural networks training

The Tensorflow 2.7.0 framework were used, each model was trained to 95% and higher accuracy on a training dataset using the Adam optimizer [53] with 0.8/0.2 train/test split. Hardware specifications: Nvidia GeForce GTX 2080Ti, Intel Core i7-10700K, 16 GB RAM.

### 5.2. Synthetic datasets

To create a synthetic dataset, a generative model Big-GAN and 10 classes of corresponding buildings from ImageNet were used: cinema, castles, church, palace, lighthouse, monastery, mosque, triumphal, mobile home, barn, each class contains 6000 images 64×64×3, (Fig. 11).

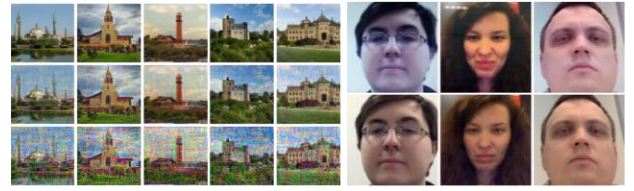

Figure 10: Left: Adversarial examples with different adversarial noise. Right: An example of screen attack [59], below are photos of people, above are spoofing attacks.

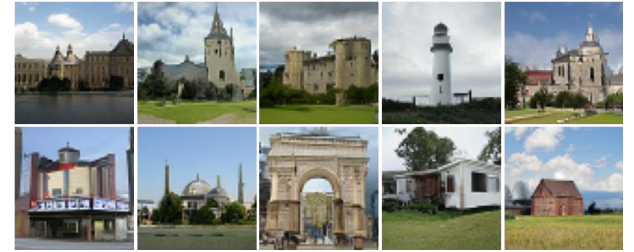

Figure 11: Synthetic dataset examples.

## 6. Conclusions and future work

As a result of our empirical research, we made an attempt to clarify, on the basis of topology and geometry, some aspects of CNNs training. We have shown that the topological properties of the data manifold are simplified in the internal representation of CNN as it goes across all depth, which is consistent with previous research. We come to the conclusion that the process of training a neural network is also associated with simplifying the topological descriptors of the data manifold, in the first epochs the topology changes insignificantly, while well-trained CNNs quickly change the data topology throughout the layer hierarchy. We have also demonstrated that the generalizability of CNN classifiers can be related to the topology of the data. Our results also include an analysis of adversarial and spoofing manifolds in terms of the topology and fractal dimension PH$_{dim}$.

To develop the topic raised in our work, we can propose to apply topology and geometry analysis for other types of CNNs architectures, as well as for recurrent models. This significantly increases the interpretability of models, which is one of the main problems of deep learning. As future studies of the properties of data manifolds, it is possible to propose the development of new characteristics, for example, a numerical analogue of the Lusternik–Schnirelmann category [63], expressing topological complexity; the topic of understanding the decisive boundaries of neural networks from the point of view of combinatorics and topology of polyhedra is also very promising.

Analyzing different aspects of deep learning from an algebraic topology perspective can improve understanding of certain phenomena or provide answers to open questions.

APPENDIX A

Estimation of persistent homological fractal dimension ($PH_{dim}$) is based on the analysis of asymptotic behavior of random minimal spanning trees and persistent homology of higher dimensions. $PH_{dim}$ is equal to the Hausdorff dimension if the measure is consistent with the regularity hypothesis.

Definition 1 [39]: A probability measure µ supported on a metric space X is d-Ahlfors regular if there exist positive real numbers c and $δ_0$ so that

$$\frac{1}{c}δ^d \leq µ(B_δ(x)) \leq cδ^d \quad (1)$$

for all x ∈ X and δ ∈ $δ_0$, where $B_δ$ is open ball of radius δ centered at x. If µ is d-Ahlfors regular on X then it is comparable to the d-dimensional Hausdorff measure on X, and the Hausdorff measure is itself d-Ahlfors regular.

According to Steele's results in [38], the length of the minimum spanning tree is a measure of the support of the distribution. It follows from this that the analysis of the asymptotic behavior of the minimal spanning tree can be applied to estimate the dimension and modeling fractals.

Theorem 1 (Steele). Let µ be a compactly supported probability measure on $\mathbb{R}^m$, m ≥ 2, and let $\{x_n\}_{n \in \mathbb{N}}$ be i.i.d. samples from µ. If 0 < α < m,

$$\lim_{n \to \infty} n^{-\frac{m-α}{m}} E_α^0(x_1, ..., x_n) \to c(α, m) \int_{\mathbb{R}^m} f(x)^{\frac{m-α}{m}} dx \quad (2)$$

with probability one, where f(x) is the probability density of the absolutely continuous part of µ, and c (α, m) is a positive constant that depends only on α and m. [80] If µ has bounded support and µ is singular with respect to Lebesgue measure, then we have that

$$\mathbb{P}\left[E_α^0(x_1, ..., x_n) = o(n^{\frac{d-α}{d}})\right] = 1$$

APPENDIX B

Evaluating the generalizing ability of CNNs through homological persistent fractal dimension in the last output layer. Figure 1 shows the relationship between $PH_{dim}$ and accuracy on a test dataset using different architectures as examples. The results of this experiment are consistent with [22]. Figure 2 shows experiments for Projected Gradient Descent (PGD) attacks.

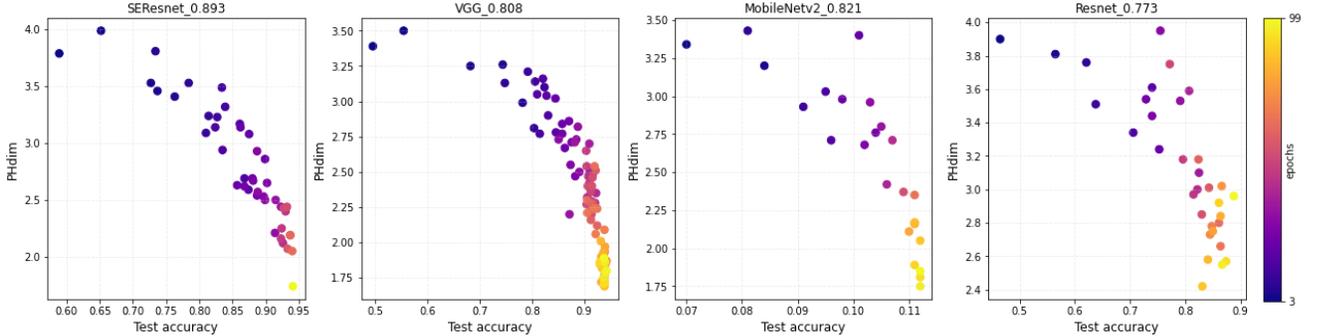

Figure 1: Relationship between $PH_{dim}$ and CNN model test accuracy at different epochs in the learning process

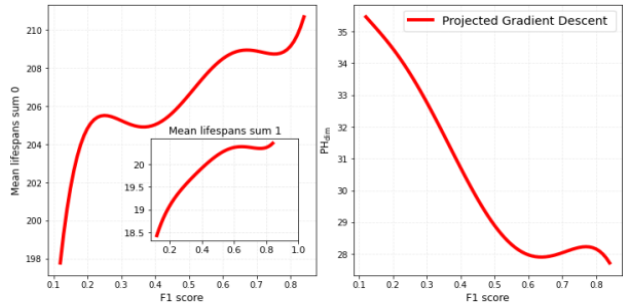

Figure 2. Left: The dynamics of topological characteristics depending on the success of the attack. Right: $PH_{dim}$.

Topological descriptors extracted from the data manifold in last layer of the CNNs for the case of higher homology dimensions can also be predictors for the generalizing ability of the neural network. Estimation of the generalizing ability of neural networks through homology of dimension 1 (Fig.3).

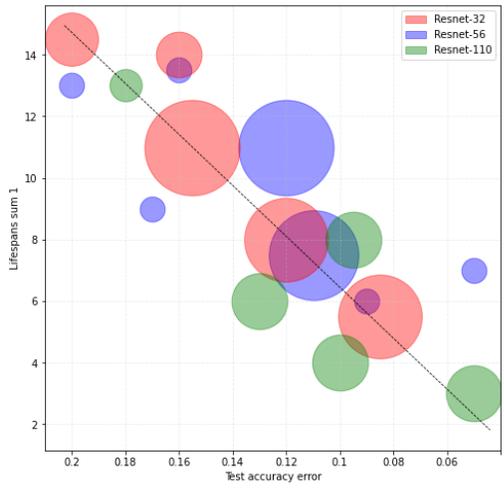

Figure 3: The relationship between model accuracy on a test dataset and topological descriptors in internal representation, 1 dimension homology case, r = -0.87.

APPENDIX C

The datasets from Table 1 are synthetic data generated with BigGAN (trained on ImageNet) by controlling intrinsic dimension through manipulation of code vectors in noise spaces [4]. Semantically, this is expressed in the diversity of the dataset (Figure 4), all images belong to the class "Castles".

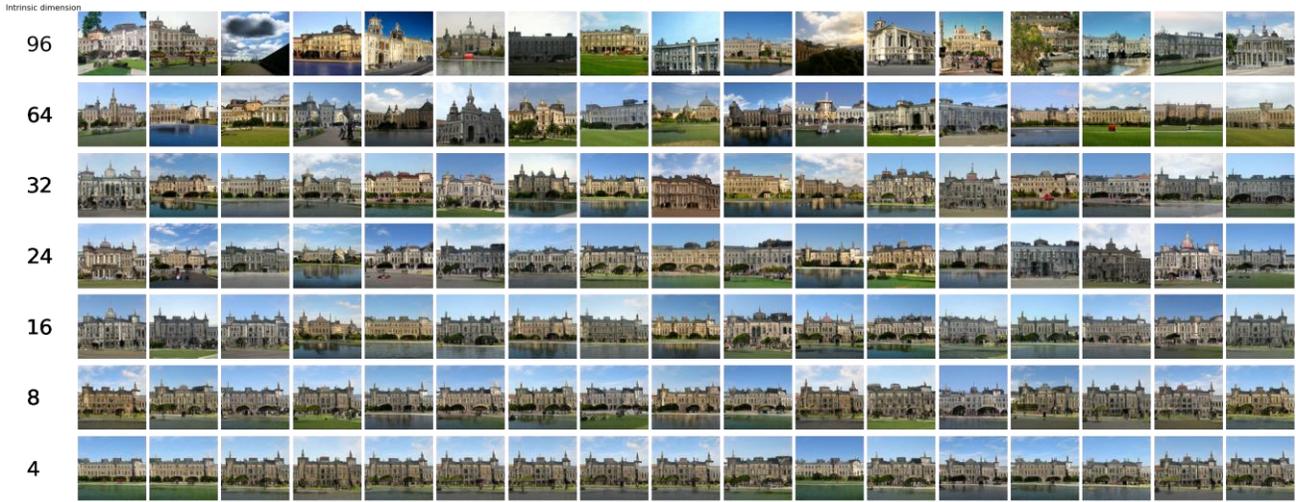

Figure 4: Relationship between $PH_{dim}$ and diversity of the dataset